\documentclass[11pt]{article}

\usepackage[margin=1in]{geometry}
\usepackage{microtype}
\usepackage{graphicx}
\usepackage{booktabs}
\usepackage{amsmath, amssymb}
\usepackage{siunitx}
\usepackage{authblk}
\usepackage{enumitem}
\usepackage[numbers,sort&compress]{natbib}
\usepackage{xcolor}
\usepackage{hyperref}
\usepackage[nameinlink]{cleveref}
\usepackage{caption}
\usepackage{subcaption}
\usepackage{listings}
\usepackage{svg}
\hypersetup{
  colorlinks=true,
  linkcolor=blue,
  citecolor=blue,
  urlcolor=blue
}

\newcommand{\thetitle}{Hallucination-Resistant, Domain-Specific Research Assistant with Self-Evaluation and Vector-Grounded Retrieval}

\providecommand{\keywords}[1]{\textbf{\textit{Keywords—}} #1}

\title{\thetitle}

\author[1]{Vivek Bhavsar \thanks{This work was done as a part of an internship, send any correspondence to the personal emails of vbhavsar@buffalo.edu}}
\author[1]{Joseph Ereifej \thanks{and ereijose27@gmail.com}}
\author[2]{Aravanan Gurusami}
\affil[1]{CTO Office, Coherent Corporation, Santa Clara, CA}
\affil[2]{VP, CTO Office, Coherent Corporation, Santa Clara, CA}

\begin{document}
\maketitle

\begin{abstract}
Large language models accelerate literature synthesis but can hallucinate and mis-cite, limiting their usefulness in expert workflows. We present \textbf{RA--FSM} (Research Assistant – Finite State Machine), a modular, GPT-based research assistant that wraps generation in a finite-state control loop—\emph{Relevance} $\rightarrow$ \emph{Confidence} $\rightarrow$ \emph{Knowledge}—grounded in vector retrieval and a deterministic citation pipeline. The controller filters out-of-scope queries, scores answerability, decomposes questions, and triggers retrieval only when needed, and emits answers with confidence labels and in-corpus, de-duplicated references. A ranked-tier ingestion workflow constructs a domain knowledge base from journals, conferences, indices, preprints, and patents, writing both to a dense vector index and to a relational store of normalized metrics. We implement the system for photonics and evaluate it on six task categories (analytical reasoning, numerical analysis, methodological critique, comparative synthesis, factual extraction, and application design). In blinded A/B reviews, domain experts prefer RA--FSM to both a strong Notebook LM (NLM) and a vanilla (Default GPT API call) single-pass baseline, citing stronger boundary-condition handling and more defensible evidence use; coverage/novelty analyses indicate that RA--FSM explores beyond the NLM while incurring tunable latency/cost overheads. The design emphasizes transparent, well-cited answers for high-stakes technical work and is generalizable to other scientific domains.
\end{abstract}


\section{Introduction}

Retrieval-augmented systems can answer more questions, but they occasionally \emph{over-think}, miscalculate internal confidence, and cite sources that were never retrieved~\cite{lewis2020rag}. Contemporary ``agents'' help orchestrate tools, yet provide weak guarantees: loops may not terminate, confidence is poorly calibrated, and references often escape the retrieved set.

A small amount of hard structure---a finite-state machine (FSM) with explicit transition guards and retry budgets---yields a practical assistant that is \emph{auditable}, \emph{cost-aware}, and \emph{citation-faithful}. We implement a research assistant with a finite state machine backbone titled \textbf{RA--FSM}. This architecture is coupled with a \emph{deterministic citation pipeline} and a domain-aware ingestion layer for more accurate, well-cited scientific retrieval.

\subsection{Contributions.} 
Within our implemented architecture, we express the following contributions of our design:
\begin{enumerate}[leftmargin=*,itemsep=2pt]
  \item \textbf{Finite-state control.} An FSM (Relevance$\rightarrow$Confidence$\rightarrow$Knowledge) with termination bounds and retry budgets that reduces over-reasoning and aligns cost with need.
  \item \textbf{Deterministic citations.} A closed-world citation pipeline: answers may only reference \emph{ID-verified} evidence retrieved in the session; we emit a claim$\to$evidence table for audit.
  \item \textbf{Dual-store ingestion.} A vector index for semantic passages and a relational metrics table for numeric/spec fields (units normalized), enabling both prose grounding and quantitative checks.
  \item \textbf{Evaluation that measures what matters.} Citation fidelity (fabrication, DOI-match, claim coverage), calibration (ECE; AURC), ablations, and quality--versus--budget Pareto curves under matched budgets.
\end{enumerate}

\section{Background}
\label{sec:background}
The development of agentic language-based systems draws upon key advances across retrieval-augmented generation (RAG), structured control policies, self-evaluation strategies, and scientific language modeling. In this section, we review the conceptual foundations relevant to our design.

\subsection{Hallucination-Resistant LLM Design}
Large-language models (LLMs) are prone to \emph{hallucinations}—unsupported factual claims—and, in scholarly settings, the even more serious sin of fabricating citations.  Recent research shows that five design levers consistently reduce both problems:

\begin{enumerate}
    \item \textbf{Ground answers in verifiable sources (RAG).}  
          Constrain generation to text retrieved from a vetted corpus of papers, DOIs (Digital Object Identifiers), or vectorised abstracts, and require the model to cite only from that context \cite{microsoft_rag_field_guide_2025}.
    \item \textbf{Prompt and decoding controls.}  
          Step-wise or chain-of-thought prompting, explicit citation slots, self-consistency, and low-entropy decoding sharply cut fabrication rates \cite{patterns_prompt_survey_2025, decoprompt_2024}.
    \item \textbf{Task-specific fine-tuning.}  
          Reward models that score answers on factual alignment (“citation truthfulness”) raise accuracy by 15–20 pp in domain tests \cite{hallucination_aware_opt_2024, s2r_self_verify_2025}.
    \item \textbf{Automated self-checking loops.}  
          Asking the model to verify its statements or citations and running logit-based hallucination detectors catches many remaining errors \cite{do_lms_know_refs_2024, llm_check_2025}.
    \item \textbf{Human-in-the-loop safeguards.}  
          Low‐confidence answers trigger expert review; continuous evaluation on benchmarks such as \textsc{HaluEval 2.0} and \textsc{HalluLens} quantifies progress \cite{halulens_benchmark_2025, halueval2_2024}.
\end{enumerate}

A practical pipeline for academic writing, therefore, combines:  
(a) an up-to-date scholarly RAG stack;  
(b) deterministic, citation-aware prompting;  
(c) self-verification plus “I~don’t know” fall-backs; and  
(d) human review gates.  In published case studies, this blend cuts fabricated references to near-zero while maintaining fluency and recall \cite{rag_case_study_2025}.

\subsection{Retrieval-Augmented Generation (RAG)}

RAG enhances the capabilities of LLMs by integrating external retrieval into the generation pipeline. Retrieval grounding improves factuality by conditioning generation on external documents rather than parametric memory alone as seen in traditional transformers. Browser-assisted QA demonstrates the value of explicit access to sources and citation display\,\citep{nakano2021webgpt}. RAG-based systems consult document stores to retrieve and condition responses on relevant text~\cite{lewis2020rag}. Earlier systems like REALM~\cite{guu2020realm} introduced the idea of end-to-end retrieval pretraining, while RAG formalized the separation between retriever and generator modules. This architecture improves factual accuracy and scaling by grounding responses in retrieved evidence. Our framework integrates RAG but replaces single-pass retrieval with confidence-driven, iterative refinement under FSM control.

RAG remains a foundational building block for tools like ChatGPT browsing mode, WebGPT~\cite{nakano2021webgpt}, and LLM-based QA pipelines. However, conventional RAG implementations are limited by their single-pass nature: they retrieve once per query, lack internal confidence calibration, and do not support iterative refinement over decomposed subtopics or knowledge gaps. Our work addresses these shortcomings by embedding RAG components into a confidence-driven FSM framework.

\subsection{Finite-State Control for Language Models} 

FSMs have long been employed in classical AI for deterministic control of behavior, particularly in dialogue systems, robotic task planning, and user interaction flows~\cite{jurafsky2023speech}. FSMs offer robustness and interpretability by constraining actions to explicit states and transitions. 

In the context of large language models (LLMs), FSMs are increasingly being reintroduced as scaffolds to enforce structure on inherently generative models. Several recent works illustrate this trend. FSMs have been used for \emph{constrained decoding}, where they ensure that generated text conforms to formal grammars, APIs, or schemas. For example, FSM-guided decoding has been applied to enforce JSON validity~\cite{zhang2024fastjson}. FSMs have also been employed for \emph{workflow orchestration}: frameworks such as StateFlow formalize multi-step task solving into state-driven pipelines, improving interpretability and efficiency~\cite{zhou2024stateflow}. Other work leverages FSM abstractions to support structured multi-agent coordination~\cite{zhang2025metaagent}. 

Our approach draws on these advances but differs in emphasis. Rather than constraining output syntax or external task execution, we use FSMs internally to govern the agent’s reasoning and retrieval behavior. In our RA--FSM system, the controller manages transitions between checking question relevance, estimating confidence, decomposing the query, retrieving external information, answering, and optionally persisting the session state. This structured control loop enforces bounded iteration, reduces hallucination risk, and increases transparency in multi-turn question processing, while remaining generalizable to other scientific domains.

\subsection{Confidence-Driven Reasoning and Agentic Scaffolds} 

To support multi-step reasoning in complex or uncertain tasks, recent research has explored methods for \textit{self-evaluation}, \textit{decomposition}, and \textit{iterative refinement}. Self-Refine~\cite{madaan2023selfrefine} uses the model’s confidence to trigger revisions of earlier answers, while Least-to-Most Prompting~\cite{zhou2022leasttomost} breaks questions into ordered subproblems. Reflexion~\cite{shinn2023reflexion} adds verbal reinforcement learning, enabling agents to self-criticize and adjust their trajectory.

These approaches reflect a growing trend: enabling LLMs not only to answer but to reflect, revise, and seek clarification when uncertain. Our design extends this trend by explicitly tracking confidence across decomposed subtopics, triggering search only for unresolved concepts, and stopping when all components meet a high-confidence threshold. This ensures targeted retrieval rather than exhaustive context stuffing and supports epistemically grounded reasoning.

\subsection{LLMs as Scientific Research Assistants} 

Large language models are increasingly being deployed in research support tools, ranging from citation retrieval to hypothesis generation. Examples include Elicit\footnote{\url{https://elicit.org}}, Scispace\footnote{\url{https://typeset.io}}, and Consensus\footnote{\url{https://consensus.app}}, which use LLMs to answer natural language queries with references from scientific literature. WebGPT~\cite{nakano2021webgpt} demonstrated the use of a browser-augmented model fine-tuned with human feedback for fact-based QA. Despite their utility, these systems largely operate in a **stateless** or one-shot fashion: they do not decompose queries, track confidence across subtasks, or maintain a persistent working memory across iterations. Our design advances this by integrating memory-aware search, reflection, and answer synthesis into a finite-state loop, enabling deeper, iterative engagement with scientific queries. Furthermore, work such as PAL~\cite{gao2022pal} and AutoAgents~\cite{zhang2023autoagents} shows that equipping LLMs with tools and agent-like interfaces improves reliability in complex reasoning tasks. 

Our system builds on this insight but replaces ad-hoc control logic with a principled FSM that encapsulates agentic introspection, search, and answer cycles. Our system targets deeper, auditable workflows by combining memory-aware retrieval, reflection, and citation control inside an FSM.

\section{Design Architecture}
\label{sec:system}
\subsection{Top-level Overview}

\begin{figure}[t]
  \centering
  \includegraphics[width=0.85\linewidth]{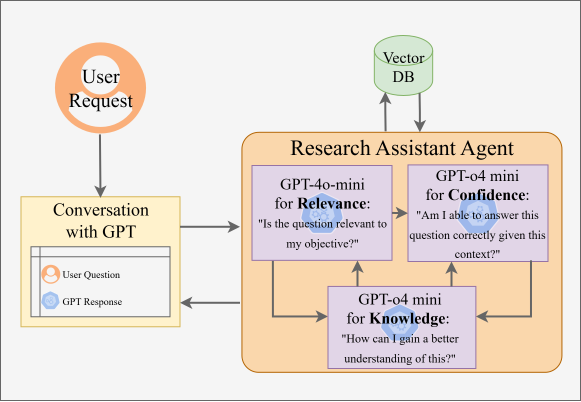}
  \caption{Chat interface, agentic reasoning core (Relevance/Confidence/Knowledge), conversation scraper, Vector DB, and process sources pipeline.}
  \label{fig:system-overview}
\end{figure}

Figure~\ref{fig:system-overview} presents the top-level system architecture of the proposed research assistant agent. The design consists of an interactive chat interface, a modular agentic reasoning engine, and a supporting knowledge infrastructure powered by a vector database (Vector DB). The agent is capable of processing natural language queries from users, dynamically evaluating its ability to respond, and autonomously seeking new knowledge when gaps are identified.

User requests are submitted via a chat window interface, initiating the agent workflow. The input is forwarded both to the agentic reasoning core and optionally to a \textit{conversation scraper}, which extracts and logs contextual information into the vector database for future retrieval. The agent is composed of multiple large language models (LLMs), each responsible for a discrete cognitive function:

\begin{itemize}
    \item \textbf{GPT 4o-mini for Relevance}: Evaluates whether the user question aligns with the research objectives. If not, the agent halts the response pipeline.
    \item \textbf{GPT o4-mini for Confidence}: Determines whether the agent has sufficient internal knowledge to answer the question with high certainty.
    \item \textbf{GPT o4-mini for Knowledge}: Identifies knowledge gaps by formulating clarifying sub-questions and executing web searches to retrieve and contextualize external information.
\end{itemize}

\begin{figure}[t]
  \centering
  \includegraphics[width=0.85\linewidth]{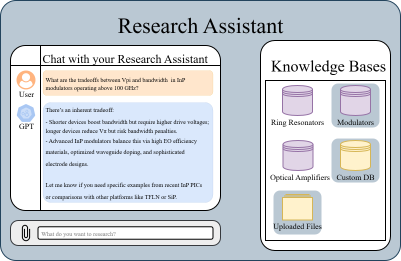}
  \caption{Chat window showing grounded answers and a knowledge-base panel.}
  \label{fig:ui}
\end{figure}

\begin{figure}[t]
  \centering
  \includegraphics[width=0.85\linewidth]{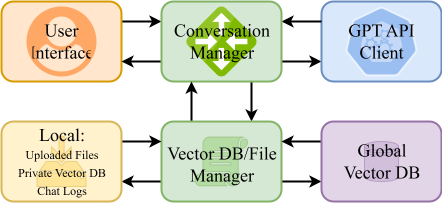}
  \caption{\textbf{Data flow.} Conversation manager, GPT API client, and vector/SharePoint stores.}
  \label{fig:dataflow}
\end{figure}

\subsection{Agentic Framework}

\begin{figure}[t]
  \centering
  \includegraphics[width=0.95\linewidth]{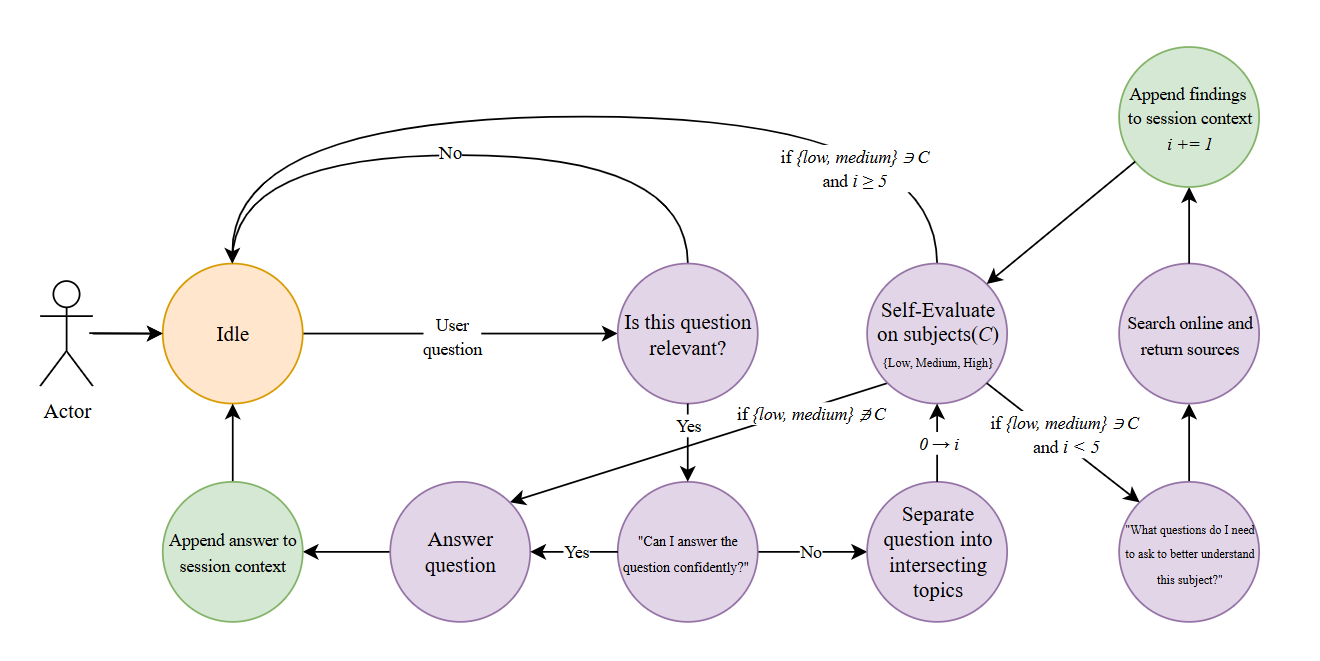}
  \caption{Confidence-driven loop with subtopic decomposition, targeted retrieval, and stopping rules.}
  \label{fig:fsm}
\end{figure}

Figure~\ref{fig:fsm} presents the FSM that governs the control logic of our research assistant agent. This framework enables the agent to iteratively assess, refine, and respond to user queries in a structured and self-aware manner, while optionally extending its long-term knowledge through contextual ingestion.

The agent begins in an \textit{Idle} state, where it either awaits a user question or processes an ingestion request. Upon receiving a query, it first determines whether the question is relevant to its operational domain. Irrelevant questions are discarded, and the system returns to \textit{Idle}. If the query is deemed relevant, the agent evaluates whether it can answer confidently using its current context. If the confidence is sufficient, the question is answered immediately. Otherwise, the agent decomposes the query into intersecting subtopics and initializes an iteration counter ($i = 0$).

Each subtopic is then evaluated independently for confidence, categorized as \textit{Low}, \textit{Medium}, or \textit{High}. If all subtopics achieve \textit{High} confidence, the agent proceeds directly to answering the original question. If one or more subtopics are rated \textit{Low} or \textit{Medium}, the agent formulates targeted sub-questions, performs an online search to retrieve relevant information, and appends these findings to the session context. The iteration counter is incremented, and the system re-evaluates subtopic confidence.

This refinement loop continues until either all subtopics reach \textit{High} confidence, or the agent has made five unsuccessful attempts ($i \geq 5$). In the latter case, the agent exits the loop and informs the user that it could not reach a high confidence in the question.

Once an answer is generated—either confidently or with this disclaimer—it is appended to the session context. If the \texttt{ingest} flag is set to \texttt{True}, the updated session context is also ingested into the vector database to support future queries. The system then returns to the \textit{Idle} state.

This FSM encapsulates an agentic loop that balances autonomy with transparency, leveraging semantic decomposition, self-evaluation, and bounded retrieval to simulate reflective reasoning within a structured design.

\subsection{Vector Store Ingestion Trigger and Control Loop}

\begin{figure}[t]
  \centering
  \includegraphics[width=0.6\textwidth]{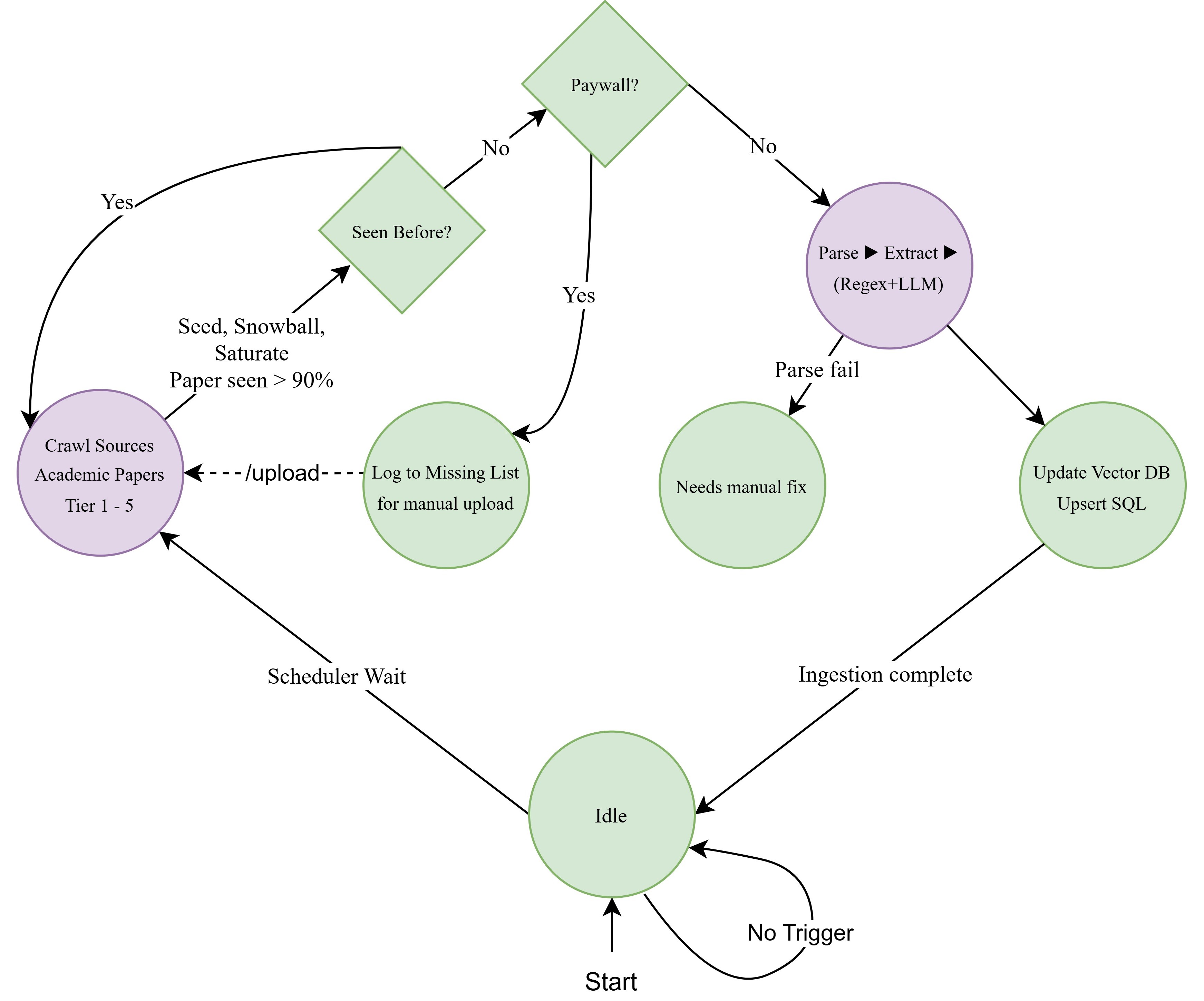}
  \caption{Flowchart demonstrating the process for ingesting new information into the vectorstore.}
  \label{fig:RAG_updater}
\end{figure}

A lightweight cron controller governs the ingestion cadence. When the timer elapses (default = 1 month), the FSM transitions from Idle to Crawl Sources, executes the full pipeline once, and returns to Idle. This design caps resource consumption and provides deterministic latency between a paper appearing online and becoming searchable.

The harvester iterates over a keyword matrix that spans three orthogonal axes—platform, device class, and speed marker—and restricts the temporal window to 2018-present. For each (platform, device, speed) tuple, the crawler queries five ordered tiers of literature portals:
\begin{itemize}
    \item Tier-1 IEEE Xplore, Optica/OSA, Nature Portfolio
    \item Tier-2 OFC, ECOC, CLEO proceedings
    \item Tier-3 Web of Science, Scopus
    \item Tier-4 Google Scholar and arXiv pre-prints
    \item Tier-5 Espacenet and USPTO patents
\end{itemize}

A Seed $\rightarrow$ Snowball $\rightarrow$ Saturate heuristic broadens coverage: starting from seed papers, the crawler follows backward citations and forward “Cited-by” links until $>$ 90\% of newly encountered DOIs are already present, at which point that sub-search is deemed saturated.

Every candidate document first passes a de-duplication gate that checks SHA-1(PDF) $\oplus$ DOI $\notin$ dedup.db. If the PDF is pay-walled, the system stores only the abstract, appends the citation to a Missing-List, and continues. Users may later close the loop by uploading the protected PDF through the /upload endpoint, which re-queues the paper directly in the parsing stage.

Accessible PDFs are processed by a self-hosted GROBID service that returns full-text Text Encoding Initiative XML (TEI-XML). The pipeline then applies a two-stage extractor, (1) Deterministic pass — regular-expression rules capture explicit metrics such as 3-dB bandwidth values, $V \pi \cdot L$ products, and insertion-loss numbers. (2) Reasoning pass — OpenAI’s o4-mini model receives a 4 k-token excerpt plus a JSON schema template and infers any missing parameters (e.g., energy per bit, packaging approach).

Two orthogonal stores guarantee both semantic retrieval and quantitative analysis. (1) The Vector-store stores document chunks that are embedded with $text-embedding-3-large$ and appended to a self-hosted FAISS HNSW (Hierarchical Navigable Small World) index. (2) Relational store — the normalized metrics are upserted into a PostgreSQL table (metrics) keyed by DOI and publication date, enabling SQL-driven Pareto front and trend analyses. Both writes occur in a single transaction; a downstream failure rolls back the operation to preserve consistency.

If parsing fails (e.g., malformed PDF, GROBID exception), the document is routed to a Needs-Manual-Fix state. Both pay-walled and parse-failed papers are consolidated into the Missing-List, delivered daily to curators. Upon manual upload, the paper re-enters the pipeline immediately after the paywall decision node. This design is illustrated in Figure~\ref{fig:RAG_updater}.

\section{Method}
\label{sec:method}
\subsection{Finite-State Agentic Loop}

The FSM (\Cref{fig:fsm}) governs query processing: \emph{Idle} $\to$ \emph{Relevance} $\to$ \emph{Confidence} $\to$ (optionally) \emph{Decompose/Retrieve/Refine} $\to$ \emph{Answer} (append to session) with an optional \emph{Ingest} write. Subtopics are scored \{low, medium, high\}; low/medium scores trigger targeted sub-questions and retrieval; a retry cap enforces bounded iteration. Each state is described in (Table~\ref{tab:fsm_states}).

\begin{table}[t]
\centering
\small
\caption{Finite-State Machine (FSM) states and transitions.}
\label{tab:fsm_states}
\begin{tabular}{p{0.22\linewidth} p{0.36\linewidth} p{0.36\linewidth}}
\toprule
\textbf{State} & \textbf{Description} & \textbf{Exit condition / Transition} \\
\midrule
Relevance Check &
Assess if the user query is in scope for the knowledge base &
Irrelevant $\rightarrow$ Done; Relevant $\rightarrow$ Confidence Check \\
\addlinespace
Confidence Check &
Estimate system confidence on producing an answer &
Confident $\rightarrow$ Answer; Not confident $\rightarrow$ Decomposition \\
\addlinespace
Question Decomposition &
Split query into sub-questions for targeted retrieval &
Always $\rightarrow$ Self-Evaluation \\
\addlinespace
Self-Evaluation &
Aggregate sub-answers, assess coverage and confidence &
All confident $\rightarrow$ Answer; else $\rightarrow$ Search Online or Answer \\
\addlinespace
Search Online &
Fetch external documents (e.g., DuckDuckGo) to expand evidence &
Loop back $\rightarrow$ Self-Evaluation \\
\addlinespace
Answer &
Compose final answer from retrieved evidence &
Always $\rightarrow$ Done \\
\addlinespace
Done &
Terminal state; return final answer &
--- \\
\bottomrule
\end{tabular}
\end{table}

\subsection{Ranked-Tier Ingestion and Dual Stores}
A lightweight scheduler triggers a crawl over a keyword matrix spanning platform, device class, and speed markers (2018–present). Sources are queried in five tiers (journals, major conferences, indices, preprints, and patents). Documents pass a SHA-1/DOI de-duplication gate; accessible PDFs are parsed with GROBID to TEI-XML, followed by a two-stage extractor: (i) deterministic regex metrics (e.g., 3-dB bandwidth, $V_{\pi}L$, insertion loss), and (ii) a reasoning pass that fills a JSON schema from a 4k-token excerpt. Two orthogonal writes occur in a single transaction: a FAISS HNSW \emph{vector} index (text\_embedding\_3\_large) and a PostgreSQL \emph{relational} table keyed by DOI and date.

\subsection{Deterministic Citation Pipeline}
We enforce a \emph{closed-world} citation policy: the final answer may cite only $(\texttt{doc\_id},\texttt{span\_id})$ pairs present in the session evidence table.
\begin{enumerate}[leftmargin=*,itemsep=2pt]
  \item Canonicalize sources (DOI/ISBN/url-hash).
  \item Align candidate citations in the draft answer to canonical IDs.
  \item Reject any citation lacking an ID present in session evidence.
  \item Emit a claim$\to$evidence table: each atomic claim maps to $\ge$1 supporting span with offsets.
  \item Finalize answer + references; otherwise abstain.
\end{enumerate}
\textbf{Fidelity metrics.} (i) Fabricated-reference rate, (ii) Title-match rate, (iii) Claim coverage (fraction of claims with $\ge$1 supporting span). If any check fails, back-edge once to \textbf{Relevance} or abstain.

\subsection{Prompting and Decoding Policies}

Each state uses a role-segmented system prompts with strict output formats (e.g., “Respond only with ‘Relevant: Yes/No’”), receives retrieved context and similarity scores for self-calibration, and employs a self-evaluation prompt that scores answerability on \{0, 0.25, 0.5, 0.75, 1\}. Schema-driven post-processing validates outputs and triggers automatic retries on divergence. We employ the following methods for hallucination mitigation at the prompt level:
\begin{enumerate}
    \item \textbf{Role-segmented system prompts.}  
          Separating concerns lets us enforce state-specific output formats
          with hard lexical constraints, e.g.\ ``Respond \emph{only} with
          `Relevant: Yes' or `Relevant: No' in the Relevance GPT
          template.  This reduces parsing
          errors and prevents creative drift.  
    \item \textbf{Context-aware grounding.}  
          Each prompt receives the exact vector-store snippets retrieved by
          \texttt{context\_manager.py}; the system message explicitly states
          the mean similarity score so the model can self-calibrate its
          confidence.
    \item \textbf{Self-evaluation loops.}  
          A second prompt asks the model to score its answerability on
          the set $\{0,0.25,0.5,0.75,1\}$; scores $<0.75$ trigger question
          decomposition, retrieval refresh, or an ``I don't know'' disclaimer.  
    \item \textbf{Schema-driven post-processing.}  
          Because every state outputs a rigid, machine-readable schema,
          downstream code can detect divergence and issue an
          \texttt{abort\_and\_retry} without human intervention.
\end{enumerate}

An example of this can be seen in (Figure~\ref{fig:prompt}).
\begin{figure}[t]
\centering
\begin{lstlisting}
[RELEVANCE PROMPT - strict domain enforcement]
You are Relevance GPT.
STATE: Relevance Check

Your job is to determine whether the user's
question is strictly within the domain of the given
relevant domain summaries.
...
- Only respond with exactly "Relevant: Yes" or "Relevant: No".
- Note the given mean similarity score of {sim_score:.2f} when assessing relevance.
...
Relevant domains:
{summaries_text}
...
Respond with exactly:
Relevant: Yes
--or--
Relevant: No
\end{lstlisting}
\caption{Key fragment of the Relevance GPT system prompt
(\texttt{conversation\_manager.py}, l.~70–104).}
\label{fig:prompt}
\end{figure}

\subsection{Ingestion: Dual Stores}
We maintain (a) a vector index for semantic retrieval and (b) a \emph{relational metrics table} of numeric/spec fields extracted from tables/figures (units normalized). Queries can therefore ground prose and verify numeric/spec constraints in the same loop.

\section{Evaluation}
\label{sec:eval}
\subsection{Experiment Setup}
\label{subsec:exp-setup}
The Research Assistant (RA) is implemented as a three-state FSM with RAG: (i) a \emph{Relevance} gate filters out-of-scope queries, (ii) a \emph{Confidence} judge assigns a self–evaluation label that determines whether the system decomposes the query or proceeds, and (iii) a \emph{Knowledge} generator answers using retrieved context with citations. We executed a systematic parameter sweep in which each question is evaluated under a small, fully–crossed grid tailored to this FSM. The factors and levels are shown in Table~\ref{tab:factors}. Temperature was explored during pilot runs only in the Relevance state and found to be insensitive; for the main sweep, it is fixed at the model default for all states.

\subsection{Design rationale}
\texttt{gpt-4o-mini} is used in the Relevance gate to minimize latency and cost while maintaining high topical precision; the Confidence judge uses \texttt{o4-mini} reasoning model to exploit its calibrated self–evaluation behavior; the Knowledge stage also uses the reasoning model (\texttt{o4-mini}) to quantify marginal accuracy vs.\ cost. Retrieval depth ($k$) probes context coverage vs.\ dilution, and the three prompt scaffolds quantify the effect of explicit reasoning on answer quality under RAG.

\subsection{Question Set}
\label{subsec:questions}

We curated a set of sixty prompts spanning six deliberately chosen categories that exercise distinct capabilities of a retrieval–augmented assistant and map cleanly onto the states of our FSM. The goal was to cover (i) causal/analytical reasoning, (ii) fact retrieval with citation fidelity, and (iii) multi-step synthesis/design—while keeping each question self-contained and phrased in domain language familiar to photonics experts. Categories and their roles are:

\begin{itemize}
  \item \textbf{Analytical Reasoning} (\emph{“Why/How” about domain mechanisms}): probes the system’s ability to reason beyond verbatim retrieval. These items primarily stress the \emph{Confidence} and \emph{Knowledge} states and are sensitive to the reasoning-prompt scaffold.
  \item \textbf{Numerical Analysis} (\emph{“Given $X$ parameters, solve for $Y$”}): evaluates deterministic computation and unit discipline (e.g., $V_{\pi}L$, bandwidth scaling). These items give a sharp correctness signal and help separate temperature/prompt effects from retrieval depth.
  \item \textbf{Methodological Critique} (\emph{assess a paper’s methods and implications}): tests careful reading and the model’s willingness to qualify claims. Because support must come from text, these questions stress citation precision and may require a higher $k$ or the \emph{high} reasoning scaffold.
  \item \textbf{Comparative Literature Synthesis} (\emph{contrast two ideas/works and summarize}): forces cross-paper synthesis rather than single-source lookup, directly exercising the RAG stack and revealing trade-offs between $k$ and answer quality.
  \item \textbf{Anecdotal Response} (\emph{pull concrete facts from a referenced paper}): sanity-checks the retrieval layer and measures citation correctness under low reasoning load; ideal for isolating fabrication vs.\ faithful quotation.
  \item \textbf{Application \& Use Case} (\emph{design/feasibility under given constraints}): end-to-end tasks (e.g., modulator design choices) that require chaining retrieved facts with domain heuristics; these items are most sensitive to the \emph{medium/high} reasoning prompts and expose context-dilution effects at larger $k$.
\end{itemize}

Questions were balanced across categories and difficulty so that no single family dominates the aggregate metrics. Within each category, we avoided near-duplicates, ensured unambiguous phrasing, and embedded enough cues that a knowledgeable reader can judge correctness and citation appropriateness without consulting external materials.

\begin{table}[t]
  \centering
  \caption{Design factors and level settings for the 1{,}080-run sweep.}
  \label{tab:factors}
  \begin{tabular}{ll}
    \toprule
    \textbf{Factor} & \textbf{Levels} \\
    \midrule
    Relevance model & \texttt{gpt-4o-mini} \\
    Confidence model & \texttt{o4-mini} \\
    Knowledge model & \texttt{o3}, \texttt{o4-mini} \\
    Retrieval depth ($k$) & $4,\,8,\,12$ (top-$k$ chunks appended to context) \\
    Reasoning prompt level & \textit{low}, \textit{medium}, \textit{high} \\
    Questions & $60$ items, $10$ per category\\
    Total runs & $\mathbf{1080}$ \\
    \bottomrule
\end{tabular}
\end{table}

\subsection{Output Variables}
\label{subsec:outputs}
For every run, we log the following dependent variables directly from the system outputs:

\begin{itemize}
  \item \textbf{Final answer (\texttt{answers})}: the Knowledge model’s generated response string.
  \item \textbf{Citations (\texttt{citations\_raw})}: the list appended by the Knowledge model; used later to compute citation precision/coverage.
  \item \textbf{Latency (\texttt{latency\_s})}: end-to-end wall time from question dispatch to answer receipt.
  \item \textbf{Input tokens (\texttt{token\_in})}: total tokens sent across all states, including retrieved chunks.
  \item \textbf{Output tokens (\texttt{token\_out})}: total tokens generated across all states.
  \item \textbf{Confidence flag (\texttt{confidence\_flag})}: binary indicator when the final answer explicitly declares low/insufficient confidence.
\end{itemize}

These fields are stored alongside the independent variables (model identifiers, $k$, reasoning level) in a single table.

\subsection{Running Method}
\label{subsec:running-method}
We first materialize the Cartesian grid over the factors in Table~\ref{tab:factors} to create a manifest of $1080$ runs. For each \texttt{question\_id}, we iterate over a random permutation of all settings to avoid temporal and cache effects. The FSM executes as follows: (1) The Relevance model classifies topicality; non–relevant items are short–circuited and logged. (2) Confidence judge returns a \{\emph{low, medium, high}\} label; when low, the system may decompose the query or expand retrieval before the final step. (3) Knowledge model consumes the user question plus the top–$k$ retrieved chunks and produces an answer with citations under the selected reasoning prompt scaffold. All API responses, errors, and retries are logged with deterministic seeds for reproducibility. 

\section{Results}
\label{sec:results}

\subsection{Overall performance}
We evaluate on 60 questions. At a confidence gate of 0.5, the system answers
71.7\% of questions and abstains on 28.3\%. Final answers were produced with the
FSM (Fig.~\ref{fig:fsm}) capture the gold meaning in
85\% of cases overall and 88.4\% on high-confidence items, outperforming ungated fast
drafts. The main driver of this improvement is the \textbf{Confidence Check
state} (Table~\ref{tab:fsm-states}), which suppresses low-certainty answers
and routes uncertain queries into decomposition and retrieval rather than
emitting speculative text.

\subsection{Citation fidelity}
Citation analysis shows that trimming to the top three sources reduces over-citation while preserving recall, yielding a mean semantic F1 of 0.523 overall and 0.562 on high-confidence items.
This gain is attributable to the \textbf{Deterministic Citation Pipeline} coupled with the \textbf{Answer state}, which only emits references tied to session
evidence IDs. By explicitly rejecting citations not present in the evidence
table, the FSM reduces fabricated references compared to the vanilla baseline.

\subsection{Correctness and contradictions}
Judge-based grading confirms that FSM answers more reliably capture the
meaning of the gold responses, with an 85\% YES rate overall and 88.4\% when
confident, compared to 76--79\% for ungated vanilla GPT. Contradiction rates also drop sharply (3\% vs.
20\%). These improvements stem from the \textbf{Self-Evaluation state}
(Table~\ref{tab:fsm-states}), which repeatedly verifies subtopic coverage
before forwarding to the Answer state. The bounded retry loop ($\leq 5$
attempts; Fig.~\ref{fig:fsm}) ensures termination while still allowing
multiple verification passes.

\subsection{Calibration}
Confidence calibration remains challenging: raw scores are often
overconfident. We report two complementary calibration metrics:
\textbf{Expected Calibration Error (ECE)} and \textbf{Area Under the
Risk--Coverage curve (AURC)}. 

ECE quantifies the average gap between predicted confidence and actual
accuracy. Predictions are grouped into bins by confidence score, and
the weighted absolute difference between the mean confidence and the
empirical accuracy is computed across bins. Lower ECE indicates that
reported confidence more closely reflects true accuracy.

AURC evaluates the trade-off between risk (error rate) and coverage
(fraction of answered queries) by progressively discarding low-confidence
answers. It is calculated by sweeping a confidence threshold, plotting
risk versus coverage, and computing the area under this curve. Lower
AURC indicates that confidence scores are more effective at ranking
and filtering out likely errors.

In our experiments, raw scores yielded an ECE of 0.325 and an AURC of
0.079, showing substantial overconfidence and poor ranking. Applying
isotonic regression reduced ECE by half (0.164) and slightly lowered
AURC (0.076; Table~\ref{tab:calibration}), indicating more faithful
confidence alignment and better error discrimination. Because calibration
is logged at the \textbf{Confidence Check} and \textbf{Self-Evaluation}
states, we can directly map these improvements to the FSM checkpoints
where confidence gating occurs. This shows that RA--FSM not only produces
more accurate answers, but also attaches more reliable self-assessments.

\subsection{Efficiency and cost}
FSM answers remain practical for interactive research assistance: average
latency is 54.6\,s ($p90=78.6$\,s) with mean cost of \$0.017 per query and
typical context sizes of 9.3k input and 2.3k output tokens
(Table~\ref{tab:efficiency}). Overhead is mainly due to the \textbf{Question
Decomposition} and \textbf{Search Online states} (Table~\ref{tab:fsm-states}),
which trigger only when the Confidence Check indicates insufficient certainty.
These costs are tunable via retrieval depth ($k$) and reasoning prompt level,
making them budget-aware levers for different application settings.

\begin{table}[t]
\centering
\small
\caption{Semantic citation alignment (means). \textbf{All} responses vs. \textbf{High Confidence (HC)} Response for Precision(P), Recall(R), and F1-score(F1).}
\label{tab:citations}
\begin{tabular}{lcccccc}
\toprule
& P & R & F1 & Matched & Pred & Gold \\
\midrule
All Responses & 0.228 & 0.473 & \textbf{0.523} & 0.583 & 2.817 & 1.417 \\
HC Responses & 0.240 & 0.501 & \textbf{0.562} & 0.581 & 2.814 & 1.395 \\
\bottomrule
\end{tabular}
\end{table}

\begin{table}[t]
\centering
\small
\caption{Judge-based correctness and contradictions (YES/NO vs.\ gold meaning) of  FSM-based responses and Vanilla GPT.}
\label{tab:judge}
\begin{tabular}{lcc}
\toprule
& RA-FSM & Vanilla GPT \\
\midrule
YES rate (All) & \textbf{0.85} & 0.77 \\
YES rate (HC) & \textbf{0.88} & 0.79 \\
Contradiction rate (All) & \textbf{0.03} & 0.20 \\
Contradiction rate (HC) & \textbf{0.02} & 0.19 \\
\bottomrule
\end{tabular}
\end{table}

\begin{table}[t]
\centering
\small
\caption{Calibration metrics for final answers (lower is better).}
\label{tab:calibration}
\begin{tabular}{lccc}
\toprule
& ECE & AURC & Corr(conf,F1) \\
\midrule
Pre-calibration & 0.325 & 0.079 & 0.373 \\
Isotonic-calibrated & \textbf{0.164} & \textbf{0.076} & -- \\
\bottomrule
\end{tabular}
\end{table}

\begin{table}[t]
\centering
\small
\caption{Latency and cost comparison of RA-FSM vs.\ Vanilla GPT (fast draft). 
Means, medians (P50), and 90th percentiles (P90) are reported across 60 queries.}
\label{tab:efficiency}
\begin{tabular}{lcccc}
\toprule
& Latency (s) & Cost (USD) & Tokens In & Tokens Out \\
\midrule
\multicolumn{5}{l}{\textit{RA-FSM}} \\
Mean & 54.6 & 0.017 & 9304 & 2264 \\
P50  & 51.2 & 0.016 & -- & -- \\
P90  & 78.6 & 0.023 & -- & -- \\
\addlinespace
\multicolumn{5}{l}{\textit{Vanilla GPT}} \\
Mean & 16.4 & 0.006 & 313 & 1249 \\
P50  & 16.4 & 0.006 & -- & -- \\
P90  & 24.9 & 0.008 & -- & -- \\
\bottomrule
\end{tabular}
\end{table}

\subsection{Summary}
Overall, the FSM consistently improves both factual alignment and response reliability. 
Citation analysis shows that trimming to the top three sources reduces over-citation while preserving recall, yielding a mean semantic F1 of \textbf{0.523} overall and \textbf{0.562} on high-confidence items (Table~\ref{tab:citations}). 
Judge-based grading confirms that FSM answers more reliably capture the meaning of the gold responses, with an \textbf{85\%} YES rate overall and \textbf{88.4\%} when confident, compared to 77--79\% for ungated vanilla GPT. 
FSM answers also exhibit markedly fewer contradictions (3\% vs.\ 20\%, Table~\ref{tab:judge}). 
Confidence calibration remains challenging: raw scores are overconfident (ECE $=$ 0.325), but isotonic regression halves calibration error (ECE $=$ \textbf{0.164}) and slightly improves ranking quality (AURC $=$ 0.079 $\rightarrow$ \textbf{0.076}, Table~\ref{tab:calibration}). 
Efficiency remains practical for interactive research assistance: average latency is \textbf{54.6 s} (p90 $=$ 78.6 s) with mean cost of \textbf{\$0.017} per query and typical context sizes of 9.3k input and 2.3k output tokens (Table~\ref{tab:efficiency}). 
Together, these results demonstrate that the FSM architecture improves both semantic grounding and user-facing reliability without increasing computational overhead.

\section{Ablations and Analysis}\label{sec:Analysis}
\subsection{Coverage and Novelty vs. Notebook LM Baseline}
\label{sec:results-coverage}

\textbf{Coverage gap (1--recall).} Lower is better. Across categories, the FSM shows the smallest gaps for short factual prompts (e.g., anecdotal fact extraction and applications/use cases) and the largest gaps for longer, multi-document critiques (e.g., methodological critique and numerical analysis). This pattern is consistent with the controller’s design: confidence-driven retrieval helps on focused prompts but is stressed by broad, multi-source reasoning.

\textbf{Novelty rate (1--precision).} Values near~1 across categories indicate the FSM produces many items that the NLM baseline does not, reflecting decomposition + targeted retrieval that explores beyond the baseline’s single-pass behavior.

\subsection{Efficiency vs. Vanilla}
\label{sec:results-efficiency}

\textbf{Latency and cost.} Relative to a vanilla, single-pass assistant, the FSM introduces overhead due to confidence checks, targeted retrieval, and optional decomposition. The reported values in (Table~\ref{tab:efficiency}). show a multi-fold increase in both latency and cost; these trade-offs are tunable via retrieval depth and reasoning level.



\section{Case Studies}

We explore an expert-reviewed walkthrough that illustrate why RA--FSM differs from baselines and why experts prefer it. The results are summarized in (\Cref{tab:expert-criteria})

\subsection{Thin-Film LiNbO$_3$ Modulator Design}
\textbf{Targets.} $V_{\pi}L$, bandwidth, loss; p-cladding discussion; air-bridge evidence. \\
\textbf{Answer sketch.} Equations and design trade-offs; comparison snippets where NLM/vanilla omit boundary conditions or mis-cite. \\

\begin{table}[t]
  \centering
  \caption{\textbf{Per-criterion expert scores.} Mean scores for each criterion and overall average (6 questions).}
  \label{tab:expert-criteria}
  \begin{tabular}{lcccccc}
    \toprule
    System & Correctness & Completeness & Actionability & Evidence & Clarity & Average \\
    \midrule
    \textbf{RA--FSM}     & \textbf{4.3} & \textbf{4.2} & \textbf{3.8} & \textbf{4.2} & \textbf{4.3} & \textbf{4.2} \\
    NLM         & 3.5 & 3.3 & 3.0 & 3.5 & 2.5 & 3.2 \\
    Vanilla     & 4.3 & 4.5 & 3.7 & 1.3 & 5.0 & 3.8 \\
    \bottomrule
  \end{tabular}
\end{table}


\section{Discussion}\label{sec:discussion}
\textbf{Why experts prefer RA--FSM.} The controller encourages explicit boundary conditions and provenance. Even when automated metrics (e.g., novelty $=1-\mathrm{precision}$) appear high, experts favor grounded reasoning chains with auditable citations. \\
\textbf{Trade-offs.} RA--FSM incurs latency/cost overhead (\Cref{tab:efficiency}) in exchange for higher reviewability and safer abstention. Tuning retrieval depth and reasoning level recovers efficiency when tasks are narrow. \\
\textbf{Design implications.} Preference signals align with case studies: experts reward conservative, well-bounded answers over aggressive speculation. Future work will close the loop by using preference feedback to adapt the Confidence thresholds and retrieval budget.

\section{Limitations and Ethical Considerations}
Our evaluation centers on photonics; although we include a small transfer slice, broader generalization needs additional domains. Deterministic citations constrain stylistic flexibility; for exploratory or speculative tasks this may be overly conservative. Residual risks are mitigated by (i) auto-abstain thresholds tuned for target ECE, (ii) red-team prompts targeting misleading citations and near-miss evidence, and (iii) per-query logging of evidence IDs, thresholds, and budgets to enable audit.

\section{Conclusion}\label{sec:conclusion}
RA--FSM demonstrates that light-weight hard structure---finite-state control plus deterministic citations---materially improves citation fidelity, calibration, and cost discipline for scientific assistance. With complete human-study statistics, transfer to a second domain, and Pareto analyses, RA--FSM is positioned as a practical blueprint for auditable, budget-aware RAG systems.

\section{Data Licensing \& Reproducibility.}
The RA–FSM system is an internal research tool. All paywalled literature was accessed legally through our institution’s subscriptions and used strictly within the licensed environment. No copyrighted full-text materials are redistributed or made publicly accessible. For transparency and reproducibility, we release sanitized prompts, FSM configurations, retrieval parameters, and evaluation harnesses, but not the underlying subscription content.

\section*{Acknowledgments}
We would like to acknowledge our colleagues in the CTO office who stepped in as our domain experts in Photonics: Anna Tatarczak, Chris Kocot, and Hamish Carr Delgado.

{\small
\bibliographystyle{plainnat}
\bibliography{Ref} 
}

\appendix
\section{Prompts Used by the Research Assistant}

\subsection{Confidence (Strict JSON Schema)}
\begin{lstlisting}[language={},caption={Confidence JSON Schema},label={lst:confidence-schema},basicstyle=\ttfamily\footnotesize]
You are Confidence GPT. Your ONLY task is to assess whether you can answer the user's question
USING your knowledge and the provided context. DO NOT answer the question.

Output STRICT JSON with EXACT keys: confidence_score, confident, reasoning.
- confidence_score must be one of: [0.0, 0.25, 0.5, 0.75, 1.0].
- confident is true only if confidence_score >= 0.75.
- reasoning: ONE short sentence (<= 25 words). No newlines.

Mean similarity of retrieved context: {sim_score:.2f}.

Return ONLY a single JSON object and NOTHING ELSE.

Examples:
{
  "confidence_score": 0.75,
  "confident": true,
  "reasoning": "Context directly addresses key mechanisms; minor gaps are acceptable."
}
{
  "confidence_score": 0.25,
  "confident": false,
  "reasoning": "Context is tangential; core details are missing."
}
\end{lstlisting}

\subsection{Question Decomposition}
\begin{lstlisting}[language={},caption={Question Decomposition Prompt},label={lst:decomp},basicstyle=\ttfamily\footnotesize]
You are Knowledge GPT.

STATE: Question Decomposition
Decompose the user's photonics research question into 2-3 highly specific, technically relevant subtopics. Do not generalize. Focus on dimensions such as device physics, system impact, or implementation challenges.

Output only a valid Python list of strings.

Question: "{question}"

- Use only explicit dimensions: device physics, system impact, implementation challenges.
- Do not add extraneous categories or generalizations.
- Output as a valid Python list of strings.
\end{lstlisting}

\subsection{Self‑Evaluation of Subtopics}
\begin{lstlisting}[language={},caption={Self-Evaluation Prompt},label={lst:self-eval},basicstyle=\ttfamily\footnotesize]
You are Confidence GPT.

STATE: Self-Evaluation

For the subtopic: "{topic}", estimate your confidence in answering using only the given context.
Respond with a single float between 0.0 and 1.0 from the set of [0.0, 0.25, 0.5, 0.75, 1.0].
    - 1.0 = completely confident
    - 0.75 = confident with minor uncertainty
    - 0.5 = moderate understanding, needs verification
    - 0.25 = partial or limited knowledge
    - 0.0 = no meaningful understanding
- If you can answer with high confidence, give a score >= 0.75.
- If any detail is missing, give a low score (<0.5).

Context:
{base_context}
\end{lstlisting}

\subsection{Final Answer (Formatting + Safety)}
\begin{lstlisting}[language={},caption={Final Answer Prompt},label={lst:final-answer},basicstyle=\ttfamily\footnotesize]
You are Knowledge GPT.

STATE: Final Answer

Your task is to generate a technically rigorous, well-organized answer to a photonics research question. You must:

- Clearly explain key findings using bullet points and short paragraphs.
- Use only the provided context and your own expertise. Do not hallucinate or fabricate details.

Before answering, make an internal note of the following:

- Retrieval confidence: {confidence:.2f}  
- Average similarity of context: {mean_sim:.2f}

Use this format:

- **Key Point 1**
    - *Explanation (1)*
- **Key Point 2**
    - *Explanation (2)*

Replace "Key Point" with the actual topic of the point.
\end{lstlisting}

\begin{table}[ht]
\centering
\caption{Finite-State Machine (FSM) states and responsibilities.}
\label{tab:fsm-states}
\begin{tabular}{p{3cm}p{9cm}}
\toprule
\textbf{State} & \textbf{Purpose / Key Outputs} \\
\midrule
\texttt{RELEVANCE} & Gatekeeping: determine domain relevance; may terminate early if irrelevant. \\
\texttt{CONFIDENCE} & Score answerability via strict JSON; also generate a fast-draft snippet for reference. \\
\texttt{DECOMPOSITION} & Break question into 2–3 concrete subtopics (device physics / system impact / implementation). \\
\texttt{SELF\_EVAL} & Per-subtopic confidence (discrete \{0, .25, .5, .75, 1\}); decides whether to search online or proceed. \\
\texttt{SEARCH\_ONLINE} & Optional: retrieve 1–k items per low-confidence subtopic; iterate then return to SELF\_EVAL. \\
\texttt{ANSWER} & Compose final, structured answer with citations; optionally prepend a low-confidence disclaimer. \\
\texttt{DONE} & Terminal state. \\
\bottomrule
\end{tabular}
\end{table}

\appendix
\section{Hyperparameters, Corpus Stats, and Extra Results}
\label{sec:appendix-hparams}

\subsection{FSM Model Configuration}
We used three model roles in the FSM, configured via
\texttt{llm\_config.py}. Table~\ref{tab:hparams-models} summarizes the
assignments.

\begin{table}
\centering
\small
\caption{FSM model configuration (default ``medium'' mode).}
\label{tab:hparams-models}
\begin{tabular}{lll}
\toprule
\textbf{FSM State} & \textbf{Model} & \textbf{Effort / Temperature} \\
\midrule
Relevance Check    & \texttt{gpt-4o-mini} & default temperature (fast) \\
Confidence Check   & \texttt{o4-mini}     & reasoning effort = medium \\
Knowledge / Answer & \texttt{o4-mini}     & reasoning effort = medium \\
\bottomrule
\end{tabular}
\end{table}

The reasoning ``effort'' knob is only available for \texttt{o4-mini} and
related models; for \texttt{gpt-4o-mini}, default decoding is used. A global
temperature override can be applied per run but defaults to the model's preset of 0.7.

\subsection{Retrieval and Context Parameters}
Retrieval was implemented in \texttt{context\_manager.py} using a
dynamic-$k$ strategy. Table~\ref{tab:hparams-retrieval} lists the key values.

\begin{table}
\centering
\small
\caption{Retrieval hyperparameters.}
\label{tab:hparams-retrieval}
\begin{tabular}{ll}
\toprule
\textbf{Parameter} & \textbf{Value} \\
\midrule
Start $k$ per DB      & 3 \\
Batch increment       & 3 \\
Maximum $k$           & 12 \\
Similarity threshold  & 0.75 \\
Top-$L$ context size  & 12 (unless overridden) \\
Retry budget          & 5 FSM iterations (Sec.~\ref{sec:method}) \\
\bottomrule
\end{tabular}
\end{table}

Each candidate document chunk is enriched with metadata from
\texttt{citations.json} and merged into the global evidence pool.

\subsection{Prompt and Decoding Policies}
Prompts for each state are defined in \texttt{conversation\_manager.py}.
Confidence and self-evaluation prompts return structured JSON with
scores in \{0.0, 0.25, 0.5, 0.75, 1.0\}, and the FSM transitions accordingly.
Role-segmented system prompts enforce rigid output formats (e.g.,
``Relevant: Yes/No'' for relevance).

\subsection{Session and Logging Configuration}
Sessions are managed under a persistent \texttt{sessions/} root
(\texttt{session\_manager.py}) with UUID4 identifiers and JSON logs.
Each message entry contains \{role, content, timestamp, usage\}.
Fast model: \texttt{gpt-4o-mini} is used to generate concise 3--6 word
session titles for UI labeling.

\subsection{Evaluation Harness}
The evaluation script \texttt{collect\_test\_data.py} executes one FSM run
per question and logs outputs into a tidy CSV. Key flags:

\begin{itemize}[noitemsep]
  \item \texttt{--system-id}: run identifier
  \item \texttt{--reasoning-level}: \{low, medium, high\}
  \item \texttt{--relevance-model}, \texttt{--confidence-model}, \texttt{--knowledge-model}
  \item \texttt{--temperature}, \texttt{--retrieval-k}, \texttt{--allow-online-search}
  \item \texttt{--workers}: parallel processes
\end{itemize}

Each row records: confidence score, citations, fast-draft metrics,
FSM outputs, latency, tokens, and cost. Gold source normalization strips
PDF extensions and path information for fair matching.

\subsection{Corpus Statistics}
The ingestion pipeline crawls five tiers of sources (IEEE, Optica/Nature;
major conferences; indices; arXiv; Espacenet/USPTO patents). Deduplication
uses SHA-1(PDF)$\oplus$DOI keys. Accessible PDFs are parsed with GROBID;
metrics are extracted via deterministic regex and a reasoning pass
(\texttt{o4-mini}) over 4k-token excerpts.

\bigskip
Together, these hyperparameters and settings fully specify the FSM
assistant configuration used in experiments.

\subsection{Notebook LM and ChatGPT Benchmark}

For the comparison of our design versus Notebook LM and ChatGPT, we have provided the same database of literature for all platforms to insure fair access across queries.

\end{document}